# Estimating Total Search Space Size
# for Specific Piece Sets in Chess

*Azlan Iqbal*[1]

ABSTRACT

Automatic chess problem or puzzle composition typically involves generating and testing various different positions, sometimes using particular piece sets. Once a position has been generated, it is then usually tested for positional legality based on the game rules. However, it is useful to be able to estimate what the search space size for particular piece combinations is to begin with. So if a desirable chess problem was successfully generated by examining 'merely' 100,000 or so positions in a theoretical search space of about 100 billion, this would imply the composing approach used was quite viable and perhaps even impressive. In this article, I explain a method of calculating the size of this search space using a combinatorics and permutations approach. While the mathematics itself may already be established, a precise method and justification of applying it with regard to the chessboard and chess pieces has not been documented, to the best of our knowledge. Additionally, the method could serve as a useful starting point for further estimations of search space size which filter out positions for legality and rotation, depending on how the automatic composer is allowed to place pieces on the board (because this affects its total search space size).

## 1    INTRODUCTION

Automatic chess problem composition is a relatively unexplored area of artificial intelligence compared to actual chess-playing. The primary reason for this is likely that playing chess can be done quite well using a combination of well-designed heuristics and relatively high computational processing speed and memory. Chess problem compositions, however, require what we humans might call 'creativity' and this is not as straightforward to implement computationally. Much of the most recent research in this particular area is actually based on my own work in relation to the Digital Synaptic Neural Substrate (DSNS) technology which is not relevant in any way to this article. Interested readers may refer to (Iqbal et al., 2016) for more information regarding the DSNS and section 2.4 of (Iqbal, 2008) for a review of notable earlier automatic chess problem composing approaches. A significant issue in automatic chess problem compositions that I have found in recent times is estimating the total search space size when composing using a particular piece set. It is generally well-known or accepted that the number of positions possible using a particular set of chess pieces grows somewhat exponentially the more pieces are used.

This is why endgame tablebases of seven pieces can contain up to 500 trillion positions or more (Lomonosov Tablebases, 2012). However, this includes *all* piece combinations of seven pieces, and not just *particular* piece sets of say, KQRBvkqr. When automatically composing a chess problem or puzzle using a particular piece set, it is useful to know in advance what the estimated search space will be for two main reasons. First, whether such a composition is likely to occur in any reasonable period of time and second, just how much 'effort' did the composing approach use to succeed, assuming it does eventually. So if approximately one million positions was all that needed to be examined in a search space of about one billion in order to successfully compose a particular type of chess problem (e.g. mate in 3 with three convention[2] filters), that would suggest that the composing approach used was quite successful or 'creative'. In the following section, I detail the mathematics behind estimating such a search space size. A simple example of the logic using fruits is first provided to help the uninitiated in this field. This is then followed by actual chess test cases. The article concludes with a summary of the main points and directions for further work.

## 2    METHODOLOGY

In the field of mathematical combinatorics, the number of combinations possible using a particular set of objects can be calculated. We will use a simple example of fruits and fruit baskets to first illustrate the concept. Let us assume we have three fruits, i.e. a banana, an apple and an orange, and six empty baskets in which we could put them. The fruits are analogous to the chess piece types and the empty baskets the empty squares on the board (imagine all sixty-four stretched out in a single line). Therefore, the idea is that any of the three fruits could be placed in any of the six different baskets just as any of the chess pieces could be placed on any of the sixty-four squares. The legality of the position is not relevant here because we only need an estimation of the number of positions that the computer will need to examine during the composing process. Legality can only be determined afterwards and is a secondary problem that does not diminish the size of the original search space.

---

[1] College of Computer Science and Information Technology, Universiti Tenaga Nasional, Kampus Putrajaya, Jalan IKRAM-UNITEN, 43000 Kajang, Selangor, Malaysia. Email: azlan@uniten.edu.my

[2] A chess problem convention may include, for example, 'no check' or 'no capture' in the first move.

So the number of possible ways to order three fruits into six baskets would be calculated using the baskets as the focus rather than the fruits. There are six empty baskets in which the banana could be placed, followed by five empty baskets for the apple and four baskets for the orange, i.e. 6 x 5 x 4 = 120 unique combinations. However, what if we had two bananas, three apples and an orange to be placed in eight empty baskets? This would be analogous to having more than one chess piece of a particular type, which is certainly possible as well. Then the calculation would be: 8 x 7 x 6 x 5 x 4 x 3 (six fruits in total) / (3! x 2!) (combinations for apples and bananas) = 1,680 unique combinations. This logic can therefore be stretched to any combination of chess pieces on sixty-four empty squares.

## 2.1 Chess Test Cases

If we take the example of say, four knights versus queen, which means KNNNNvkq (seven pieces), the calculation using the method just explained would be: 64 x 63 x 62 x 61x 60 x 59 x 58 / 4! = 130,455,400,320 unique combinations. This seems like a plausible number given that the size of the seven-piece endgame tablebase is over 500 trillion positions. This particular combination therefore represents only around 0.025% of that total. If we use the example of four knights versus two rooks, which means KNNNNvKRR (eight pieces), we must bear in mind that an endgame tablebase of eight pieces does not yet exist and would be exponentially larger than 500 trillion positions. So being able to calculate the search space of this particular combination of pieces would be useful using the method just described. The calculation is therefore: 64 x 63 x 62 x 61 x 60 x 59 x 58 x 57 / (4! x 2!) = 3,717,978,909,000 unique combinations.

Incidentally, such a position was indeed composed using the automatic chess problem composer, Chesthetica (Friedel, 2017). Three conventions were also applied, namely, no cooks, no checks and no captures in the first or key move. Clearly finding not only a legal forced mate but also one that abides by all said conventions is more difficult than just any position in the universe of nearly 4 trillion positions. Regardless, the number of positions that needed to be examined by Chesthetica (using the DSNS approach) was around 120,000 or less than 0.00000323% of the search space. This provides some useful perspective and scale about how effective the automatic composing approach is and perhaps how many such compositions could, in principle, be composed in a given amount of time. A 'systematic' brute force approach of composing would therefore clearly be futile even for this one particular combination of pieces.

## 3 CONCLUSIONS

The mathematical approach proposed in this article can be used to calculate the total number of possible positions given a set of chess pieces. The main reason for its applicability is that legality of a position is typically determined only *after* the pieces are 'randomly' placed on the board by the program. The number of such positions that could theoretically be generated and tested would therefore be useful for the programmer or experimenter to know. Further work in this area might include developing other methods to determine the raw position counts for particular piece combinations so they can be compared with the method proposed. Additionally, future automatic chess problem or puzzle composers can also be compared against each other by benchmarking their ability to compose based on these piece set search space sizes, as per the example given in section 2.1. This is especially useful using piece sets of eight or more, which do not have existing endgame tablebases and perhaps never will.

**REFERENCES**

Iqbal, M. A. B. M. (2008). *A Discrete Computational Aesthetics Model for a Zero-sum Perfect Information Game*, Ph.D. Thesis, Faculty of Computer Science and Information Technology, University of Malaya, Kuala Lumpur, Malaysia. https://goo.gl/zg6hVA

Iqbal, A., Guid, M., Colton, S., Krivec, J., Azman, S., and Haghighi, B. (2016). *The Digital Synaptic Neural Substrate: A New Approach to Computational Creativity*, 1st Edition, SpringerBriefs in Cognitive Computation, Springer International

Publishing, Switzerland. eBook ISBN 978-3-319-28079-0, Softcover ISBN 978-3-319-28078-3, Series ISSN 2212-6023. DOI: 10.1007/978-3-319-28079-0. Shorter preprint: http://arxiv.org/abs/1507.07058

Lomonosov Tablebases (2012). *About Lomonosov Tablebases*. http://tb7.chessok.com/about-site

Friedel, F. (2017). *Four Knights vs Queen Challenge II*, ChessBase News, Hamburg, Germany, 2 October. http://en.chessbase.com/post/four-knights-vs-queen-challenge-2